# DV3+HED+: A DCNNs-based Framework to Monitor Temporary Works and ESAs in Railway Construction Project Using VHR Satellite Images

**Rui Guo[1,2], Ronghua Liu[4], Na Li [3, *] Wei Liu[1,2]**

1. Aerospace Information Research Institute, Chinese Academy of Sciences, No.9 Dengzhuang South Road, Haidian District, Beijing, P.R. China 100094, guorui@radi.ac.cn (R. G.); liuwei@radi.ac.cn (W. L.);
2. Institute of Remote Sensing and Digital Earth, Chinese Academy of Sciences, No.9 Dengzhuang South Road, Haidian District, Beijing, P.R. China 100094.
3. Ping An Technology (Shenzhen) Co., Ltd, Shenzhen, P.R. China 200240; LINA420@pingan.com.cn.
4. China Institute of Water Resources and Hydropower Research, Beijing, P.R. China 100038, liurh@iwhr.com.
* Correspondence: LINA420@pingan.com.cn; Tel.: +86-10-8217-8151

**Abstract:** Current VHR(Very High Resolution) satellite images enable the detailed monitoring of the earth and can capture the ongoing works of railway construction. In this paper, we present an integrated framework applied to monitoring the railway construction in China, using QuickBird, GF-2 and Google Earth VHR satellite images. We also construct a novel DCNNs-based (Deep Convolutional Neural Networks) semantic segmentation network to label the temporary works such as borrow & spoil area, camp, beam yard and ESAs(Environmental Sensitive Areas) such as resident houses throughout the whole railway construction project using VHR satellite images. In addition, we employ HED edge detection sub-network to refine the boundary details and attention cross entropy loss function to fit the sample class disequilibrium problem. Our semantic segmentation network is trained on 572 VHR true color images, and tested on the 15 QuickBird true color images along Ruichang-Jiujiang railway during 2015-2017. The experiment results show that compared with the existing state-of-the-art approach, our approach has obvious improvements with an overall accuracy of more than 80%.

## 1. Introduction

Railways have been vital in supporting the society, people's livelihood and economic development in China over the past 40 years. The rapid development of the railway construction provides convenient transportation for people and accelerates economic and social development, but it occupies and destroys a certain amount of land resources inevitably as well. How to control and reduce the negative effects to environment brought by railway construction has become a key issue that both the administrative management and project construction department must confront and resolve.

The environment monitoring during railway construction project is to supervise and inspect the execution of environmental protection measures on the basis of the design and environment evaluation report of this project and to affirm the achievements, find out existing problems and give suggestions on countermeasures. According to the different functions, the construction project of railway consists of three parts, which are permanent works, temporary works and ESAs. The permanent works mainly contain roadbeds, tracks, stations, bridges, piers, culverts, water supply and sewerage work, and electrification facilities etc., which should be strictly checked and accepted according to the project plan during the construction. The temporary works mainly contain borrow areas, spoil areas, camps and beam yards which play a distinctly subsidiary role but have significant influences to the environment during the project construction. The ESAs mainly refer to resident houses which concern with the critical relocation affairs of nearby residents. In this paper, we mainly focus on the monitoring of temporary works and ESAs, which are illustrated in Table 1.

**Table 1.** Description of temporary works and ESAs

| Construction name | Construction type | Geometry type | Description |
|---|---|---|---|
| Borrow area | Temporary work | Polygon | An area designated as the excavation site for geologic resources, such as rock/basalt, sand, gravel, or soil. |
| Spoil area | Temporary work | Polygon | An area used to refer to material removed when digging a foundation, tunnel, or other large excavation. |
| Resident house | ESA | Polygon | Residents within 30 meters apart from the boundary of the land used by the project should be relocated. |
| Camp | Temporary work | Polygon | Camp for workers. |
| Beam yard | Temporary work | Polygon | - |

As long linear construction projects, many railways go through regions of complex terrain, which poses great difficulties to monitoring current status of temporary works and ESAs. With the advantages of low cost, periodic data acquiring, and historical data archiving, VHR satellite images are very suitable for monitoring the changes along the railway. Pixel-wise classification such as support vector machines[1], neural networks[2], random forest[3] are widely used to classify low spatial resolution (10–30m) images. In the past 10 years, GEOBIA(Geographic Object-Based Image Analysis) has been explored to deal with high spatial variability in the VHR images[4]. However, the performance of GEOBIA is inherently dependent on the level of the segmentation results. Considering the complexity of land cover that contains vegetation, water, soil and other physical land features, it is still challenging for GEOBIA to improve classification accuracy in VHR images.

From the other perspective, traditional pixel-wise classification and GEOBIA focused feature extraction approaches such as SIFT(Scale-Invariant Feature Transform)[5] and HOG(Histogram of Oriented Gradient)[6] and supervised learning algorithms. However, the two steps mentioned above are typically treated as independent approaches. DCNNs fuse the them into one network that learns semantic features at different scales and computes the score of each class at the end of the network. In recent years, DCNNs have performed quite well in computer vision tasks, such as image classification, targets detection and semantic segmentation.

In this paper, we present a novel framework for temporary works and ESAs monitoring of railway construction project using VHR satellite images. Compared to existing studies, our novel contributions are as follows:

- We construct a novel workflow for temporary works and ESAs monitoring of railway construction project using VHR satellite images.

- We construct an efficient supervised learning model for VHR images classification based on the fusion of the state of art semantic segmentation network DV3+(DeepLabV3 plus)[7] and HED (Holistically-nested Edge Detection)[8].
- For the sake of solving class imbalance problem of training data, we introduce attention loss function to the ground object boundary detection in HED.

**2. Related work**

According to the published studies, satellite remote sensing has not been used in temporary works or ESAs monitoring for railway construction project. However, satellite images had been used for monitoring the changes of light rail transport construction in Kuala Lumpur, Malaysia[9]. Likewise, Giannico[10] present site detection and EIA(Environmental Impact Assessment) method due to the construction using satellite images. Lin[11] employed UAV to monitor abandoned dreg fields of high-speed railway construction. Chang[12] detected the railway subsidiaries using interferometric synthetic aperture radar techniques. Arastounia[13] presented an automated recognition method of railroad infrastructure in rural areas using LIDAR data.

Over the last few years, methods based on FCNs(Fully Convolutional Networks) [14] have demonstrated significant improvement on PASCAL15] and MS-COCO [16] segmentation benchmarks than the traditional pixel-wise classification and GEOBIA. SegNet[17] introduced an encoder and decoder network into the pooling indices. U-Net[18] adds skip connections from the encoder features to the corresponding decoder activations. RefineNet[19] combined rough high-level semantic features and fine-grained low-level features. Inspired by SegNet and ResNet[20], LinkNet[21] introduced residual blocks to the network architecture, which made efficient use of scarce resources available on embedded platforms without any significant increase in number of parameters. PSPNet[22] concatenated the regular CNN layers and the upsampled pyramid pooling layers, carrying both local and global context information to the image. BiSeNet[23] designed spatial path and context path, and tried to use a new method to keep both spatial context and spatial detail at the same time. The Deeplab series which contains LargeFOV(DeepLab Large Field-Of-View)[24], ASPP(DeepLab Atrous Spatial Pyramid Pooling)[25], DV3(Deeplab V3)[26] and DV3+[7] employed atrous convolution, fully connected CRFs to localize the segment boundaries and encoder-decoder framework achieved a higher accuracy than previous methods.

In remote sensing research, Penatti[27] showed that a pre-trained CNN used to recognize natural image objects generalizes well to remote sensing images by transfer learning. Based on FCN, many frameworks were derived to learn features at different scales and fusing such features in many ways[28,29,30,31]. Marmanis[32] extracted scale-dependent class boundaries before each pooling level, with the class boundaries fused into the final multi-scale boundary prediction. Guo [33] extracted bounding boxes of potential ground objects which augmented the training dataset before training the DCNNs. Tian[34] presented DFCNet(Dense Fusion Classmate Network) which was jointly trained with auxiliary road dataset properly compensates the lack of mid-level information. Li[35] proposed Y-Net which contained a two-arm feature extraction module and a fusion module for road segmentation.

Nevertheless, the tendency of the DCNNs is to extract and fuse the global semantic features and local features from different scales.

**3. Satellite data collection and processing**

*3.1. Satellite data collection*

As deep learning is a data-driven method, DCNNs rely on diversity and quality of the datasets to achieve a satisfactory training accuracy and capability of generalization. Therefore, label data making for temporary works and ESAs became a critical task in the whole framework.

As shown in Table 2, we collected three types of VHR satellite data source including QuickBird, GF-2 and Google Earth images for training data labeling which mainly covered more than six recent built railway lines in China. Multi-temporal GF-2 images, 0.8m spatial resolution panchromatic images and 3.2m resolution multispectral images including red, green, blue, and infrared bands, were acquired as the main data source for ground truth labeling, training and testing. QuickBird images with 0.61m panchromatic and 2.44m multispectral spatial resolution bands, Google Earth images with RGB bands were collected as the auxiliary data source for ground truth labeling and model training.

In addition to the VHR satellite images labeling, the ESRI Shapefile datum of the railway construction project was considered as the reference data.

Table 2. VHR images summary for training and testing

| Railway Name | Quick Bird | GF-2 | Google Earth Images |
|---|---|---|---|
| Inner Mongolia-Jiangxi railway | 10 | 28 | 66 |
| Hangzhou–Changsha high-speed railway | 17 | 40 | 47 |
| Shijiazhuang–Jinan high-speed railway | 9 | 34 | 65 |
| Jiujiang–Quzhou high-speed railway | 18 | 31 | 43 |
| Sichuan–Tibet railway | 19 | 28 | 32 |
| Others | 13 | 21 | 51 |

As the construction stage of a railway could last serval years and due to the adverse effects of satellite imaging by cloud and fog, we proposed semi-annually or annually remote sensing monitoring of the target objects according to the capability of VHR satellite revisiting to the same place. In this study, we managed to collect 15 scenes of QuickBird images of the study area for model testing. The corresponding dates are from 2015 to 2017. Table 3 summarizes the scene details of the testing Ruichang-Jiujiang QuickBird images. Figure 1 illustrates the different visual appearance of the target temporary works and ESAs in VHR images.

Table 3. Ruichang-Jiujiang QuickBird images scene details

| Year | QuickBird Scenes |
|---|---|
| 2015 | 2015-03-02, 2015-03-02, 2015-04-11, 2015-04-11,2015-06-06 |
| 2016 | 2016-11-05, 2016-11-05, 2016-12-11, 2016-12-11, 2016-12-11 |
| 2017 | 2017-12-26, 2017-12-26, 2017-08-28, 2017-08-28, 2017-08-28 |

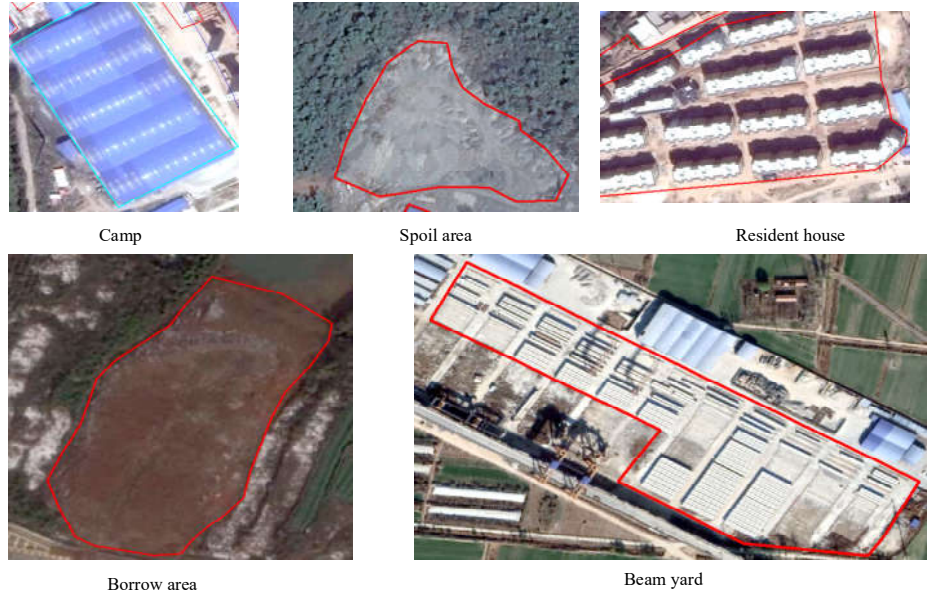

**Figure 1.** Visual appearance of the target temporary works and ESAs in VHR images

*3.2. Data processing and labeling*

Time series of QuickBird, GF-2 and Google Earth satellite images were used for training and testing in this study. The data processing workflow was shown in Figure 2. First, panchromatic and multispectral bands of QuickBird and GF-2 images were geometrically corrected as well as orthorectified. Then, we employed pansharpening method which combined the high resolution of panchromatic images with the lower resolution of multispectral ones. The advantage of such method is to get as a final result a colored image of a certain area with a high resolution, optimizing the starting panchromatic one. Last, we mosaic the VHR images with the same imaging time in the same railway construction project. Moreover, the histograms of the images were adjusted to enhance the contrast. We also employed the Google Earth RGB images as the additional data source. We search the locations of railway construction line according to its coordinates and exported the images from the Google Earth software. All processing for this study was completed using ArcGIS 10.4.1 by ESRI ©. Table 4 illustrates the ground truth sample count of target temporary works and ESAs

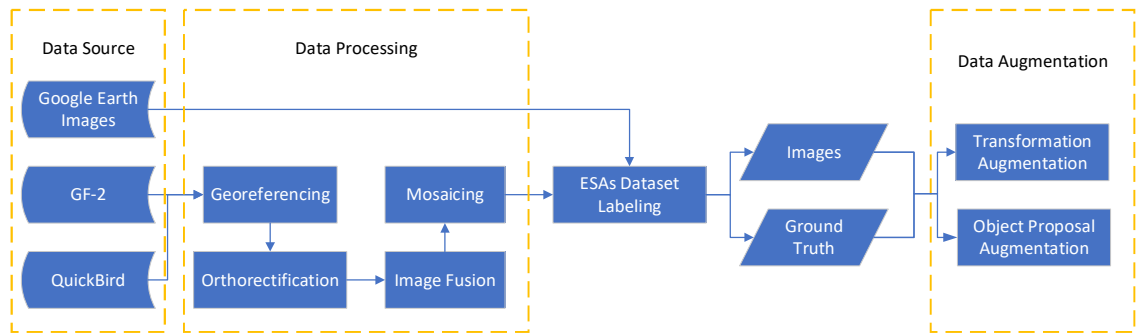

**Figure 2.** Data processing workflow

**Table 4.** Ground truth sample count of target temporary works and ESAs

| Name | Ground Truth Count | Type |
| --- | --- | --- |
| Borrow area | 106 | ESA |
| Spoil area | 294 | ESA |
| Resident houses | 1953 | ESA |

| Camp | 484 | Temporary work |
| Beam yard | 25 | ESA |

Data augmentation has been widely used for avoiding overfitting when training data are not sufficient to learn a generalizable model. In this paper, we followed the satellite data augmentation method presented by Guo[42], in which the selective search method was applied to generate bounding boxes of potential ground objects in the VHR satellite images. Thus, we obtain more valuable trained data by using unsupervised methods than simple transformation augmentation.

## 4. Methodology

DCNNs-based models and atrous convolution have proven to be the most successful methods of semantic segmentation. The VHR remote sensing images contain abundant geometric information of the ground object. In order to make better use of this information, we combine the HED boundary detection network[8] and the state-of-the-art DV3+[7] semantic segmentation network ;which integrated the advantages of ASPP and encoder-decoder for pixel-wise classification of remote sensing imagery. In addition, we employ the Attention Loss[36] to scale class-balanced cross entropy loss and upgrade the loss contribution of both false negative and false positive samples during the training process.

*4.1. Network architecture*

Our network architecture shown in Figure 5 follows the idea of Marmanis[32], which combines ground object boundary detection along with semantic segmentation. Based on VGG-16, the HED outputs a multi-scale feature map before each pooling layer for edge detection. The multi-scale feature maps are then fused into a final boundary feature map. The relation of each scale layer loss function and fusion layer loss function is illustrated as following:

$$L_{fuse} = \sum_{m=1}^{M} l_{side}^{(m)} \quad (1)$$

where $l_{side}^{(m)}$ denotes the different scale level loss function for each side output, $L_{fuse}$ denotes fusion layer loss function of the side outputs.

After boundary detection sub-network, the network fuses the original input image and the boundary prediction result and then the fusion of image and boundary are put into the DV3+ semantic segmentation sub-network. The DV3+ proposes a state-of-the-art encoder-decoder structure which employs DV3 as encoder module and a simple yet effective decoder module for natural image semantic segmentation. The DV3+ also adapts the Xception model and apply depth wise separable convolution to both encoder and decoder module, resulting in a faster and stronger network. After the processing of HED and DV3+ sub-networks, the classification results are predicted.

Due to the complexity of the ground objects and the artificial workload, we didn't label the boundary ground truth data for training manually. Considering that each ground object in label data represents a different class, the edge between different classes can be regarded as the boundary ground truth which can be produced by label data simply. In this paper, we use Sobel edge detection operator to generate the boundary ground truth data.

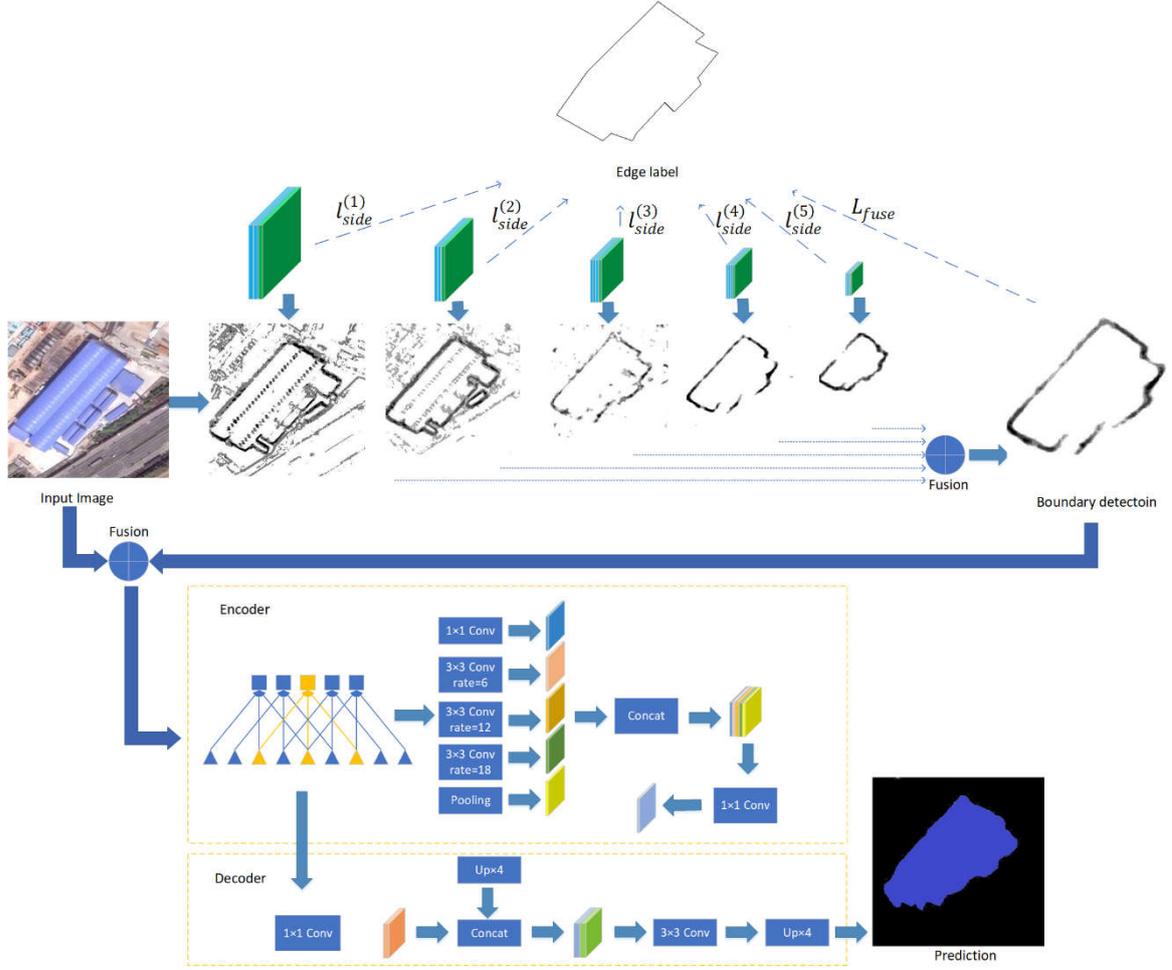

**Figure 3.** DV3+HED+ network architecture

*4.2. Attention loss function for boundary detection*

Traditionally, Cross-Entropy loss function ($L_{CE}$) is used for the training of FCN-based semantic segmentation networks. However, as the $L_{CE}$ is computed by summing overall the pixels including both foreground and background, it does not perform well for imbalanced classes. In detail, the total pixel count of boundaries is considerably much smaller than the pixel count for the entire image which contributes to numerical optimization difficulties when training neural networks. Xie[8] approach these difficulties by adopting a loss function with pixel-wise weights that automatically balance the loss between edges and non-edges. To take the imbalanced classes into account, the Class-balanced Cross-Entropy loss function ($L_{CCE}$) is defined as follows:

$$L_{CCE} = -\alpha \sum_{j \in Y_+} log(p_j) - (1-\alpha) \sum_{j \in Y_-} log(1-p_j) \qquad (2)$$

Where $p_j$ is computed on the activation value at pixel $j$ using sigmoid function, $\alpha = \frac{\overline{A_-}}{\overline{A}}$ and $1 - \alpha = \frac{\overline{A_+}}{\overline{A}}$, $\overline{A_+}$, $\overline{A_-}$ and $\overline{A}$ denote the boundary, non-boundary and the total count of a training ground truth batch, respectively. Although $L_{CCE}$ can easily classify edge pixels for common edge, it is hard to discriminate true positive and false positive samples for boundary detection under the condition of p ∈ [0.3,0.6], where most edges do not belong to boundaries of the ground objects.

The Attention Loss function ($L_{AL}$) [36] adds two adjusting parameters to the $L_{CCE}$ with $\beta > 0$ and $\gamma \geq 0$, which is defined as follows:

$$L_{AL} = -\sum_{j \in Y_+} \alpha\beta^{(1-p_j)^\gamma}\log(p_j) - \sum_{j \in Y_-}(1-\alpha)\beta^{p_j^\gamma}\log(p_j) \tag{3}$$

The parameter $\beta$ adjusts true positive and false positive loss contributions. The $L_{AL}$ penalizes misclassified samples strongly and penalizes the correctly classified samples weakly, which is more discriminating. The parameter $\gamma$ smoothly accommodate the loss on the condition of certain $\beta$ value.

**5. Experiment and results**

In this section, we describe the training settings of the experiment and present numerical and visual results. Meanwhile, we evaluate the benefits of each component of our proposed method.

*5.1. Training*

In this paper, the experiments we have done were based on the TensorFlow framework developed by Google and performed on a computer running the Ubuntu 16.04 operating system and equipped with two NVIDIA RTX 2080 Ti graphics card with 22 GB of memory. TensorFlow has been used extensively in the area of deep learning, and there are many pre-trained models that are based on it. We could finetune the models that have been validated successfully in natural image semantic segmentation.

The DCNNs-based model was trained by SGD (Stochastic Gradient Descent). To fit the model, we tiled the original images into patches of 513×513 sizes supplemented with the augmentation patches mentioned in 3.2. In every iteration, a mini-batch of patches was fed to the n for backpropagation. In all cases, a momentum of 0.9 and an L2 penalty on the network's weight decay of 0.0002 were used. The learning rate was computed dynamically between 0.007 and 1e-6. Weights were initialized following [20,37], and training ended after 50,000 iterations, when the error stabilized on the validation set.

*5.2. Classification results*

We evaluate the performance of our method based on three criteria: per-class accuracy, the overall accuracy and the average recall. The accuracy is defined as the number of true positives ($TP$) divided by the sum of the number of true positives and the number of false positives ($FP$):

$$accuracy = \frac{TP}{TP+FP} \tag{4}$$

Recall is defined as the number of true positives ($TP$), divided by the sum of the number of true positives and the number of false negatives ($FN$):

$$recall = \frac{TP}{TP+FN} \tag{5}$$

According to the presented classification workflow and network architecture, we inferred the 2km buffer VHR satellite images along the Ruichang-Jiujiang Railway line during 2015-2017 based on the trained model. The classification results are shown in Table 5.

**Table 5.** Temporary works and ESAs classification results of Ruichang-Jiujiang railway during 2015-2017

| Year | Method | Borrow-spoil area | Resident house | Camp | Beam yard | Average Recall | Overall Accuracy |
|---|---|---|---|---|---|---|---|
| 2015 | DV3+ | 78.52 | 74.37 | 81.42 | 70.84 | 77.17 | 77.23 |
|  | DV3+HED | 79.73 | 76.01 | 83.94 | 71.54 | 79.32 | 79.94 |
|  | DV3+HED+ | 80.91 | 76.56 | 84.33 | 73.18 | 79.87 | 80.05 |
| 2016 | DV3+ | 79.72 | 74.02 | 82.05 | 69.44 | 76.02 | 76.95 |
|  | DV3+HED | 80.3 | 75.87 | 83.84 | 71.03 | 78.41 | 78.72 |
|  | DV3+HED+ | 81.46 | 76.59 | 83.91 | 72.79 | 79.18 | 80.35 |
| 2017 | DV3+ | 78.28 | 75.55 | 82.96 | 65.82 | 77.3 | 78.24 |
|  | DV3+HED | 79.77 | 76.43 | 84.21 | 68.66 | 78.47 | 79.11 |
|  | DV3+HED+ | 80.53 | 76.87 | 84.73 | 69.14 | 78.53 | 80.19 |

The visual classification results are shown in Figure 6. From the perspective of target object characteristics, camps, borrow and spoil areas with more than 80% high classification accuracy are characterized by obvious features and a simple internal distribution. Among them, the camps have obvious stripe-like texture features and color features of the blue roof. The borrow and spoil areas present the color characteristics of the bare soil. Since the borrow and spoil areas produced bare soil on the ground during the construction process, which only could be distinguished accurately by the professionals. Based on the above considerations, we take the borrow and spoil areas as a same target object category. The resident houses and the beam yards are multiple mixed features. The resident houses varied widely in which buildings and bungalows coexist, mixing with small roads between the houses. There are also a small number of blue roof houses in the resident houses, which might be misclassified into camps.

From the perspective of ground truth amount, the resident houses account for the largest proportion of total ground truth samples. The beam yard proportion is the least because the number of beam yard in each railway construction is extremely small. As the DCNNs-based method is a sample-oriented classifier, the number of samples directly affects the classification accuracy.

The project construction of Ruichang-Jiujiang Railway started at June, 2014, completed at September 2017, and the main construction periods were 2015 and 2016. For the temporary works, of 2016 is the highest in the construction period, while the resident houses are increasing year by year. It is noted that after the completion of the project in 2017, the beam yard was quickly dismantled, resulting in a rapid decline in its area, while other temporary engineering features did not change significantly.

DCNNs-based workflow can be used to classify the ESAs and temporary works, the accuracy still needs to be improved. Compared with the classification results of studies on standard datasets such as ISPRS Vaihingen and Potsdam datasets, the VHR images used in this project have various qualities and spatial resolutions. Besides, the resident houses, beam yards, and camps that need to be classified were all belonged to the building category.

After the process of classification, the temporary works and ESAs is automatically prepared for change detection. However, our experience shows that trade-offs must be made between accuracy, performance and low-cost mapping. Even in the case of a very accurate automatic method, manual revision and correction of the results remain important parts of the process.

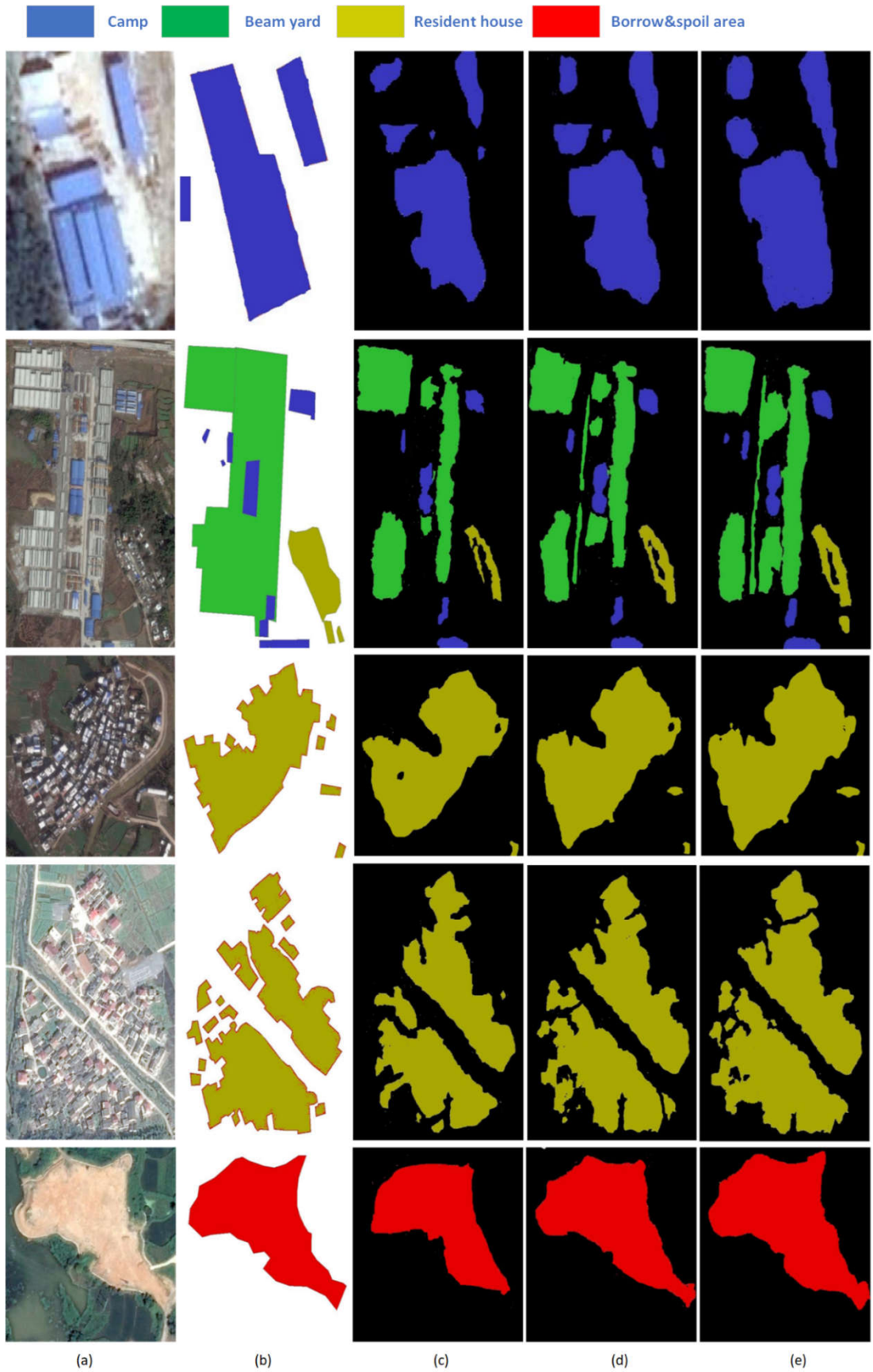

**Figure 4.** Classification result comparasion of different network architectures. (a) Original images; (b) Ground truth; (c) DV3+ classification results; (d) DV3+HED classification results;(e) DV3+HED+ classification results.

## 5.3. Effectiveness of attention loss function

As shown in Figure 5, we evaluate our model on the validating dataset while training, which presents a qualitative validating comparison between the DV3+HED with $L_{CCE}$ loss function and DV3+HED+ with $L_{AL}(\beta = 4, \gamma = 0.4)$ loss function. Figure 5(a) shows the validating total loss comparison on the validating dataset. The total loss combines cross entropy of semantic segmentation, different scale level loss for each side output and their fusion loss, and $L_2$ regular loss of each parameter in the network. For DV3+HED, the loss converges on around 26 after about 22000 iterations. Better than DV3+HED, the total loss of DV3+HED+ converges on around 22 after about 28000 iterations. Figure 5(b) shows that the validating accuracy of DV3+HED+ is superior to the DV3+HED on the validating dataset after 4000 training iterations. The validating accuracies of DV3+HED and DV3+HED+ converge on 0.75 and 0.83 respectively.

Figure 6 shows the boundary detection validating accuracy comparison of HED and HED+ sub-networks. Similar with Figure 5(b), the boundary detection validating accuracy of DV3+HED+ is lower than DV3+HED firstly, then surpass it and converge on around 0.8.

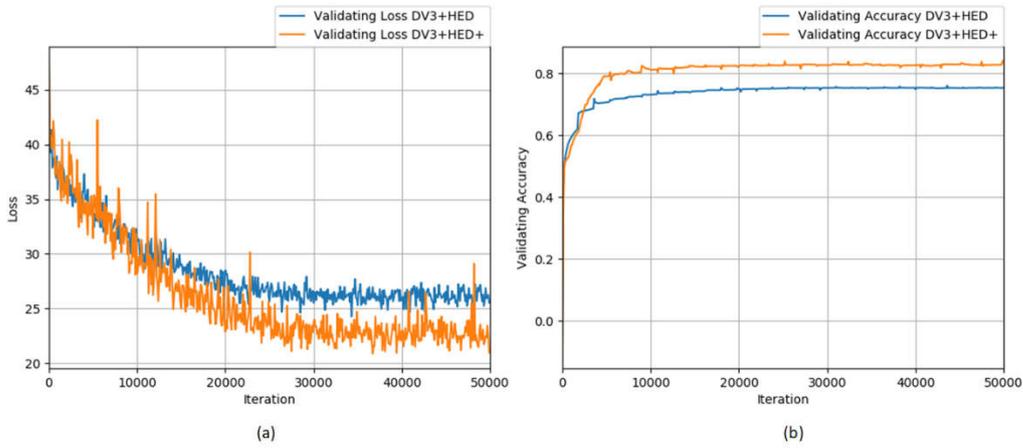

**Figure 5.** (a) Validating loss comparison of DV3+HED and DV3+HED+; (b) Validating accuracy comparison of DV3+HED and DV3+HED+.

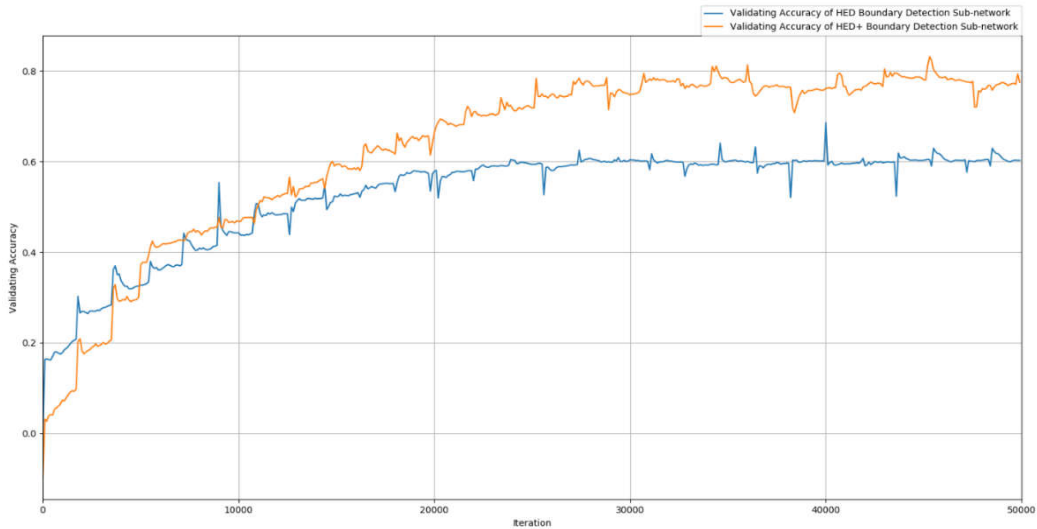

**Figure 6.** Boundary detection validating accuracy comparison of HED and HED+ sub-networks.

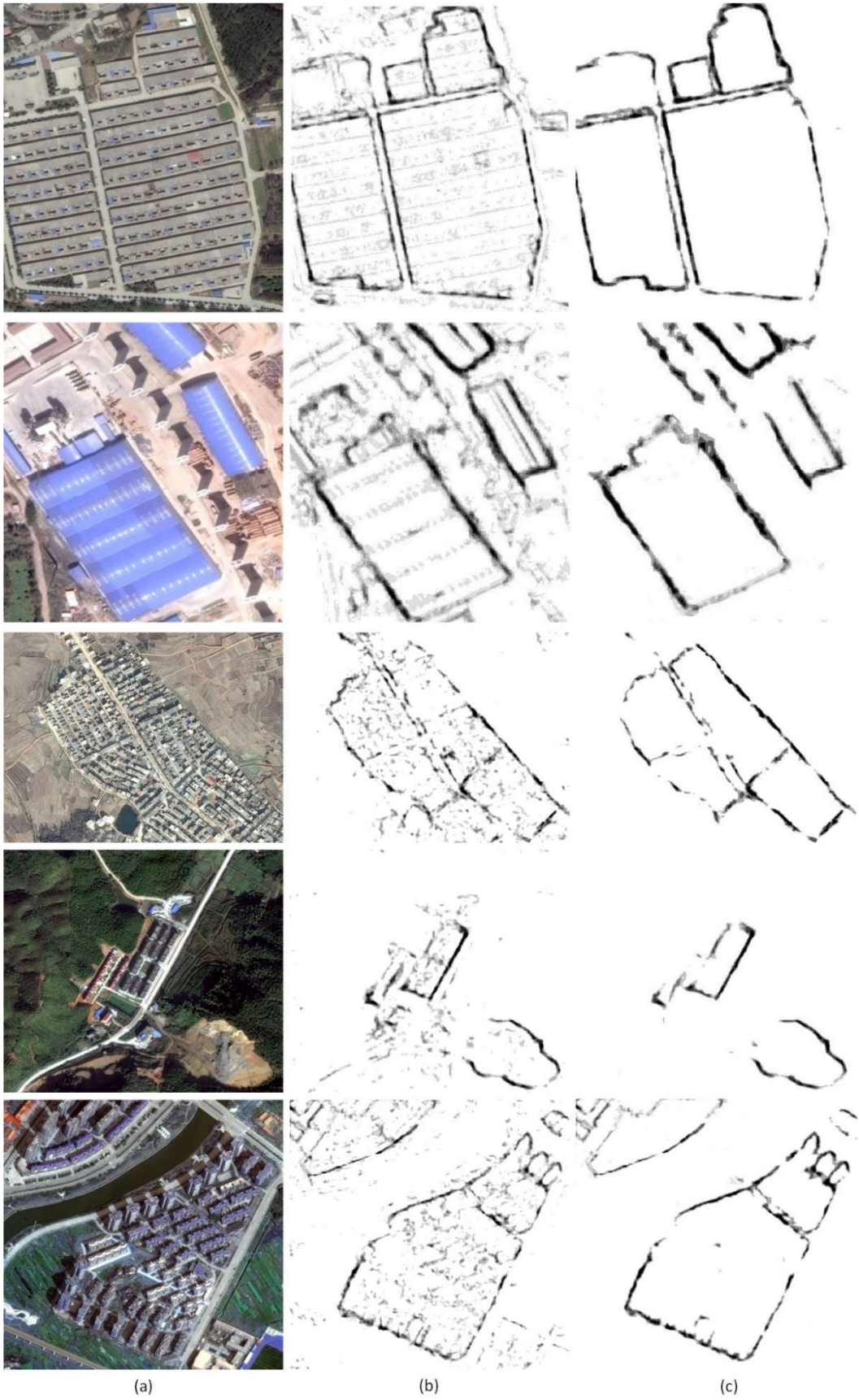

**Figure 7.** Boundary detection result comparison of HED and HED+sub-networks. (a) Original images; (b) Predicted boundaries with class-balanced cross-entropy loss function; (c) Predicted boundaries with attention loss function

Figure 7 shows boundary detection result comparison of HED and HED+ sub-networks. Although the $L_{CCE}$ loss function employed by the HED sub-network can detect common boundary pixels of the ground objects, the edge pixels inside the ground objects are also detected. It is hard for $L_{CCE}$ loss function to discriminate true positive and false positive edges where most edge pixels do not belong to boundary. Therefore, as an important part of input data to the subsequent pixel-wise classification network, the boundary detection results mixed with false positive edges is not conducive. As the boundary detection results shown in Figure 7, the $L_{AL}$ loss function employed by the HED+ sub-network puts more focus on hard, misclassified samples and classify the boundary more precise than $L_{CCE}$.

**6. Conclusions**

DCNNs-based model has been proved to be efficient in the semantic segmentation of ground objects of construction activities. To support the monitoring of temporary works and ESAs of railway construction projects, we introduced a novel DCNNs-based monitoring framework using VHR satellite images. The framework was developed and tested with Ruichang-Jiujing railway construction project in China. Focusing on classification problems for target ground objects, the proposed DV3+HED+ network labels the class of each pixel in the input images. With reference to the previous state-of-the-art semantic segmentation results, the framework detected and calculate the precise changes among multitemporal images.

The main purpose of this paper is to propose a DCNNs-based classification workflow for providing reference data to the environmental soil and water conservation supervision department to reduce the manual labor. The proposed framework has been developed into a system, which allowed the pixel-wise classification module working as a plugin. We also open the source code of the DV3+HED+ network on GitHub for researchers who are interesting with our works (https://github.com/xjock/deeplebv3plus-hedplus). In further studies, multiple source satellite images need to be considered in order to make the semantic segmentation monitoring framework more practical.

**Acknowledgments:** The work is supported by the National Key Research and Development Project under Grant No. 2018YFE010010001-3.

**Author Contributions:** Rui Guo designed and performed experiments, analyzed data and wrote the paper; Na Li supervised the research, and additionally provided comments and revised the manuscript. Ronghua Liu provided the VHR satellite images and ground truth data. Wei Liu was helpful in improving the English writing.

**Conflicts of Interest:** The authors declare no conflicts of interest